\documentclass{svproc}
\usepackage{times}

\usepackage{multicol}
\usepackage{amssymb}
\usepackage{subfig}
\usepackage{multicol}
\usepackage[bookmarks=true]{hyperref}
\usepackage{graphicx,color,subfig}
\usepackage{amsfonts, amssymb, amsmath,verbatim, multirow}
\usepackage{enumerate}
\usepackage{breqn}
\usepackage{amsmath}
\usepackage{times}
\usepackage{algorithm}
\usepackage{algorithmic}
\usepackage{easybmat}

\DeclareMathOperator*{\argmax}{arg\,max}

\newcommand{\mb}[1]{{\boldsymbol{#1}}}

\newcommand{\trsp}{{\!\scriptscriptstyle\top}}

\newcommand{\tp}[1]{\text{\tiny#1}}
\newcommand{\ty}[1]{{\scriptscriptstyle{\mathcal{#1}}}}
\newcommand{\diag}{\mathrm{diag}}

\usepackage[symbol]{footmisc}

\newcommand{\rulesep}{\unskip\ \vrule\ }

\makeatletter
\newcommand\footnoteref[1]{\protected@xdef\@thefnmark{\ref{#1}}\@footnotemark}
\makeatother

\renewcommand{\thefootnote}{\fnsymbol{footnote}}

\begin{document}
\mainmatter              
\title{Generalizing Robot Imitation Learning with Invariant \\ Hidden Semi-Markov Models}
\titlerunning{Robot Imitation Learning with Hidden Semi-Markov Models}  
%

\author{Ajay Kumar Tanwani\footnoteref{inst1}$^{,}$\footnote{\label{inst2}University of California, Berkeley.}, Jonathan Lee\footnoteref{inst2}, Brijen Thananjeyan\footnoteref{inst2}, Michael Laskey\footnoteref{inst2}, \\ Sanjay Krishnan\footnoteref{inst2}, Roy Fox\footnoteref{inst2}, Ken Goldberg\footnoteref{inst2}, Sylvain Calinon\footnote{\label{inst1}Idiap Research Institute, Switzerland.} \let\thefootnote\relax\footnote{Corresponding author: \email{ajay.tanwani@berkeley.edu}}}
\authorrunning{Tanwani \textit{et al.}} 
\tocauthor{Ajay Tanwani}
\institute{}
\maketitle              
\begin{abstract}
Generalizing manipulation skills to new situations requires extracting invariant patterns from demonstrations. For example, the robot needs to understand the demonstrations at a higher level while being invariant to the appearance of the objects, geometric aspects of objects such as its position, size, orientation and viewpoint of the observer in the demonstrations. In this paper, we propose an algorithm that learns a joint probability density function of the demonstrations with invariant formulations of hidden semi-Markov models to extract invariant segments (also termed as sub-goals or options), and smoothly follow the generated sequence of states with a linear quadratic tracking controller. The algorithm takes as input the demonstrations with respect to different coordinate systems describing virtual landmarks or objects of interest with a task-parameterized formulation, and adapt the segments according to the environmental changes in a systematic manner. We present variants of this algorithm in latent space with low-rank covariance decompositions, semi-tied covariances, and non-parametric online estimation of model parameters under small variance asymptotics; yielding considerably low sample and model complexity for acquiring new manipulation skills. The algorithm allows a Baxter robot to learn a pick-and-place task while avoiding a movable obstacle based on only $4$ kinesthetic demonstrations.

\end{abstract}
\section{Introduction}
Generative models are widely used in robot imitation learning to estimate the distribution of the data for regenerating samples from the model \cite{Argall09}. Common applications include probability density function estimation, image regeneration, dimensionality reduction and so on. The parameters of the model encode the task structure which is inferred from the demonstrations. In contrast to direct trajectory learning from demonstrations, many problems arise in robotic applications that require higher contextual level understanding of the environment. This requires learning invariant mappings in the demonstrations that can generalize across different environmental situations such as size, position, orientation of objects, and viewpoint of the observer. Recent trend in imitation leaning is forgoing such a task structure for end-to-end supervised learning which requires a large amount of training demonstrations.

\begin{figure*}[tbp]
\centering
\includegraphics[trim={2.3cm 3.0cm 3cm 4.6cm},clip,scale = 0.43]{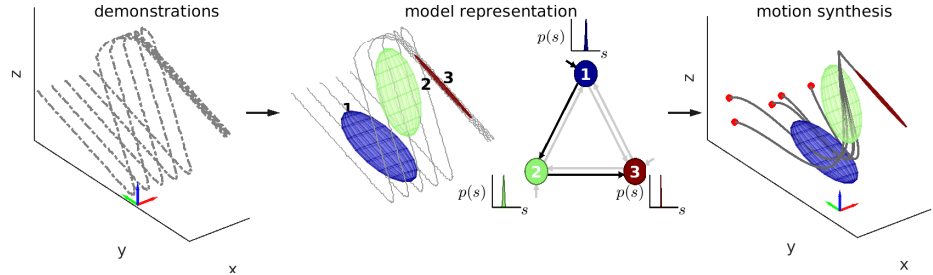}
\caption{\small Conceptual illustration of hidden semi-Markov model (HSMM) for imitation learning: \textit{(left)} $3$-dimensional Z-shaped demonstrations composed of $5$ equally spaced trajectory samples, \textit{(middle)} demonstrations are encoded with a $3$ state HMM represented by Gaussians (shown as ellipsoids) that represent the blue, green and red segments respectively. The transition graph shows a duration model (Gaussian) next to each node, \textit{(right)} the generative model is combined with linear quadratic tracking (LQT) to synthesize motion in performing robot manipulation tasks from $5$ different initial conditions marked with orange squares (see also Fig. \ref{fig: HSMMLQT}).}\label{fig: SOSC_Coceptual}
\end{figure*}

The focus of this paper is to learn the joint probability density function of the human demonstrations with a family of \textbf{Hidden Markov Models (HMMs)} in an \textbf{unsupervised} manner \cite{Rabiner89}. We combine tools from statistical machine learning and optimal control to segment the demonstrations into different components or sub-goals that are sequenced together to perform manipulation tasks in a smooth manner. We first present a simple algorithm for imitation learning that combines the decoded state sequence of a hidden semi-Markov model \cite{Rabiner89,Yu10} with a linear quadratic tracking controller to follow the demonstrated movement \cite{borrelli11}(see Fig. \ref{fig: SOSC_Coceptual}). We then augment the model with a task-parameterized formulation such that it can be systematically adapted to changing situations such as pose/size of the objects in the environment \cite{Calinon16,Tanwani16,Wilson99}. We present latent space formulations of our approach to exploit the task structure using: 1) mixture of factor analyzers decomposition of the covariance matrix \cite{McLachlan03}, 2) semi-tied covariance matrices of the mixture model \cite{Tanwani16}, and 3) Bayesian non-parametric formulation of the model with Hierarchical Dirichlet process (HDP) for online learning under small variance asymptotics \cite{Tanwani16a}. The paper unifies and extends our previous work on encoding manipulation skills in a task-adaptive manner \cite{Tanwani18,Tanwani16,Tanwani16a}. Our objective is to reduce the number of demonstrations required for learning a new task, while ensuring effective generalization in new environmental situations.
\subsection{Related Work}
Imitation learning provides a promising approach to facilitate robot learning in the most `natural' way. The main challenges in imitation learning include \cite{Nehaniv04}: 1) \textbf{what-to-learn} -- acquiring meaningful data to represent the important features of the task from demonstrations, and 2) \textbf{how-to-learn} -- learning a control policy from the features to reproduce the demonstrated behaviour. Imitation learning algorithms typically fall into \textbf{behaviour cloning} or \textbf{inverse reinforcement learning (IRL)} approaches. IRL aims to recover the unknown reward function that is being optimized in the  demonstrations, while behaviour cloning approaches directly learn from human demonstrations in a supervised manner. Prominent approaches to imitation learning include Dynamic Movement Primitives \cite{Ijspeert13}, Generative Adversarial Imitation Learning \cite{Ho16}, one-shot imitation learning \cite{Duan17} and so on \cite{Osa18}.

This paper emphasizes learning manipulation skills from human demonstrations in an unsupervised manner using a family of hidden Markov models by sequencing the atomic movement segments or
primitives. HMMs have been typically used for recognition and generation of movement skills in robotics \cite{Khokar14,Lee10a,Racca16,Yang18}. Other related application contexts in imitation learning include options framework \cite{Fox17,Krishnan17}, sequencing primitives \cite{Medina17}, and neural task programs \cite{Xu18}.

A number of variants of HMMs have been proposed to address some of its shortcomings, including: 1) how to bias learning towards models with longer self-dwelling states, 2) how to robustly estimate the parameters with high-dimensional noisy data, 3) how to adapt the model with newly observed data, and 4) how to estimate the number of states that the model should possess. For example, \cite{Kulic08} used HMMs to incrementally group whole-body motions based on their relative distance in HMM space. \cite{Lee10a} presented an iterative motion primitive refinement approach with HMMs. \cite{Niekum12} used the Beta Process Autoregressive HMM for learning from unstructured demonstrations. Figueroa et al. used the transformation invariant covariance matrix for encoding tasks with a Bayesian non-parametric HMM \cite{Figueroa17}.

In this paper, we address these shortcomings with an algorithm that learns a hidden semi-Markov model \cite{Rabiner89,Yu10} from a few human demonstrations for segmentation, recognition, and synthesis of robot manipulation tasks (see Sec. \ref{sec: HMM}). The algorithm observes the demonstrations with respect to different coordinate systems describing virtual landmarks or objects of interest, and adapts the model according to the environmental changes in a systematic manner in Sec. \ref{sec: TPHMM}. Capturing such invariant representations allows us to compactly encode the task variations than using a standard regression problem. We present variants of the algorithm in latent space to exploit the task structure in Sec. \ref{sec: LS-HMM}.  In Sec. \ref{sec: Exp}, we show the application of our approach to learning a pick-and-place task from a few demonstrations, with an outlook to our future work.
\section{Hidden Markov Models}\label{sec: HMM}
\textbf{Hidden Markov models (HMMs)} encapsulate the spatio-temporal information by augmenting a mixture model with latent states that sequentially evolve over time in the demonstrations \cite{Rabiner89}. HMM is thus defined as a doubly stochastic process, one with sequence of hidden states and another with sequence of observations/emissions. Spatio-temporal encoding with HMMs can handle movements with variable durations, recurring patterns, options in the movement, or partial/unaligned demonstrations. Without loss of generality, we will present our formulation with semi-Markov models for the remainder of the paper. Semi-Markov models relax the Markovian structure of state transitions by relying not only upon the current state but also on the duration/elapsed time in the current state, i.e., the underlying process is defined by a \textit{semi-Markov} chain with a variable duration time for each state. The state duration stay is a random integer variable that assumes values in the set $\{1, 2, \ldots, s^{\max}\}$. The value corresponds to the number of observations produced in a given state, before transitioning to the next state. \textbf{Hidden Semi-Markov Models} (HSMMs) associate an observable output distribution with each state in a semi-Markov chain \cite{Yu10}, similar to how we associated a sequence of observations with a Markov chain in a HMM. 

Let $\{\mb{\xi}_t\}_{t=1}^{T}$ denote the sequence of observations with $\mb{\xi}_t \in \mathbb{R}^{D}$ collected while demonstrating a manipulation task. The state may represent the visual observation, kinesthetic data such as the pose and the velocities of the end-effector of the human arm, haptic information, or any arbitrary features defining the task variables of the environment. The observation sequence is associated with a hidden state sequence $\{z_t\}_{t=1}^{T}$ with $z_t \in \{1 \ldots K \}$ belonging to the discrete set of $K$ cluster indices. The cluster indices correspond to different segments of the task such as reach, grasp, move etc. We want to learn the joint probability density of the observation sequence and the hidden state sequence. The transition between one segment $i$ to another segment $j$ is denoted by the transition matrix $\mb{a} \in \mathbb{R}^{K \times K}$ with $a_{i,j} \triangleq P(z_t = j | z_{t-1} = i)$. The parameters $\{\mu_j^{S}, \Sigma_j^{S}\}$ represent the mean and the standard deviation of staying $s$ consecutive time steps in state $j$ as $p(s)$ estimated by a Gaussian $\mathcal{N}(s|\mu_j^{S},\Sigma_j^{S})$. The hidden state follows a categorical distribution with $z_t \sim \text{Cat}(\mb{\pi}_{z_{t-1}})$ where $\mb{\pi}_{z_{t-1}} \in \mathbb{R}^{K}$ is the next state transition distribution over state ${z_{t-1}}$ with $\Pi_i$ as the initial probability, and the observation $\mb{\xi}_t$ is drawn from the output distribution of state $j$, described by a multivariate Gaussian with parameters $\{\mb{\mu}_j, \mb{\Sigma}_j\}$. The overall parameter set for an HSMM is defined by $\Big\{ \Pi_i, \{a_{i,m} \}_{m=1}^K, \mb{\mu}_i,\mb{\Sigma}_i, \mu_i^{S},\Sigma_i^{S} \Big\}_{i=1}^K$. 
\subsection{Encoding with HSMM} \label{sec: encodingHSMM}
For learning and inference in a HMM \cite{Rabiner89}, we make use of the intermediary variables as: 1) \textbf{forward variable}, $\alpha^{\tp{HMM}}_{t,i}\triangleq P(z_t = i, \mb{\xi}_1 \ldots \mb{\xi}_t | \theta)$: probability of a datapoint $\mb{\xi}_t$ to be in state $i$ at time step $t$ given the partial observation sequence $\{\mb{\xi}_1,\ldots,\mb{\xi}_t\}$, 2) \textbf{backward variable}, $\beta^{\tp{HMM}}_{t,i}\triangleq P(\mb{\xi}_{t+1} \ldots \mb{\xi}_T | z_t = i, \theta)$: probability of the partial observation sequence $\{\mb{\xi}_{t+1},\ldots,\mb{\xi}_T\}$ given that we are in the $i$-th state at time step $t$, 3) \textbf{smoothed node marginal} $\mb{\gamma}^{\tp{HMM}}_{t,i}\triangleq P(z_t = i | \mb{\xi}_1 \ldots \mb{\xi}_T, \theta)$: probability of $\mb{\xi}_t$ to be in state $i$ at time step $t$ given the full observation sequence $\mb{\xi}$, and 4) \textbf{smoothed edge marginal} $\zeta^{\tp{HMM}}_{t,i,j}\triangleq P(z_t = i, z_{t+1} = j | \mb{\xi}_1 \ldots \mb{\xi}_T, \theta)$: probability of $\mb{\xi}_t$ to be in state $i$ at time step $t$ and in state $j$ at time step $t+1$ given the full observation sequence $\mb{\xi}$. Parameters $\Big\{ \Pi_i, \{a_{i,m} \}_{m=1}^K, \mb{\mu}_i,\mb{\Sigma}_i \Big\}_{i=1}^K$ are estimated using the EM algorithm for HMMs, and the duration parameters $\{\mu_i^{S},\Sigma_i^{S}\}_{i=1}^{K}$ are estimated empirically from the data after training using the most likely hidden state sequence $\mb{z}_t = \{z_1 \ldots z_T \}$ (see App. \ref{app: EM-HMM} for details). 

%
\subsection{Decoding from HSMM} \label{sec: samplingHSMM}
\begin{figure}[tbp]
\centering
\includegraphics[trim={5.2cm 0.5cm 1.5cm 1cm},clip,scale=0.48]{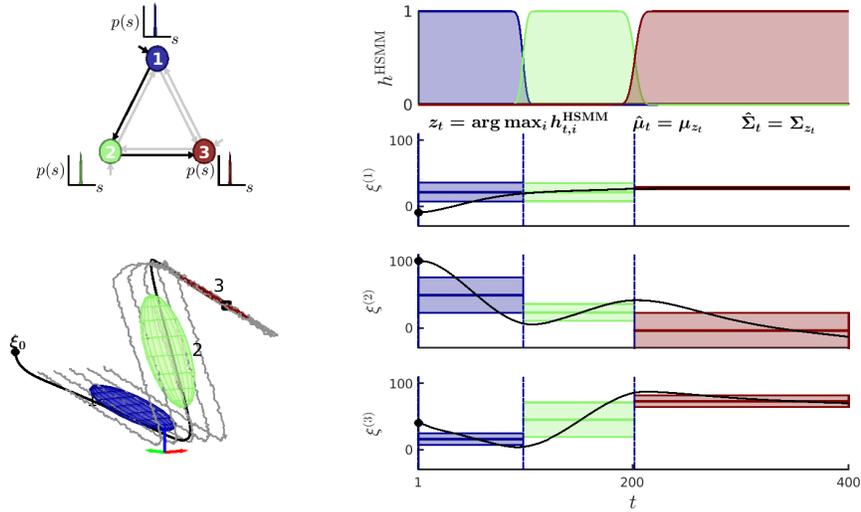}
\caption{\small Sampling from HSMM from an unseen initial state $\mb{\xi}_0$ over the next time horizon and tracking the step-wise desired sequence of states $\mathcal{N}(\mb{\hat{\mu}}_t, \mb{\hat{\Sigma}}_t)$ with a linear quadratic tracking controller. Note that this converges although $\mb{\xi}_0$ was not previously encountered.}\label{fig: HSMMLQT}
\end{figure}

Given the learned model parameters, the probability of the observed sequence $\{\mb{\xi}_1 \ldots \mb{\xi}_t\}$ to be in a hidden state $z_t = i$ at the end of the sequence (also known as \textit{filtering} problem) is computed with the help of the forward variable as
\begin{equation}
P(z_t \; | \; \mb{\xi}_1, \ldots, \mb{\xi}_t) = h_{t,i}^{\tp{HMM}} = \frac{\alpha_{t,i}^{\tp{HMM}}}{\sum_{k=1}^{K}\alpha_{t,k}^{\tp{HMM}}} = \frac{\pi_i \mathcal{N}(\mb{\xi}_t |\mb{\mu}_i, \mb{\Sigma}_i)}{ \sum_{k=1}^{K}\pi_k \mathcal{N}(\mb{\xi}_t |\mb{\mu}_k, \mb{\Sigma}_k)}.
\end{equation} 
Sampling from the model for predicting the sequence of states over the next time horizon $P(z_{t}, z_{t+1}, \ldots, z_{T_p} \; | \; \mb{\xi}_1, \ldots, \mb{\xi}_t)$ can be done in two ways: \textbf{1) stochastic sampling: }the sequence of states is sampled in a probabilistic manner given the state duration and the state transition probabilities. By stochastic sampling, motions that contain different options and do not evolve only on a single path can also be represented. Starting from the initial state $z_t = i$, the $s$ duration steps are sampled from $\{\mu_i^{S}, \Sigma_i^{S}\}$, after which the next transition state is sampled $z_{t+s+1} \sim \mb{\pi}_{z_{t+s}}$. The procedure is repeated for the given time horizon in a receding horizon manner; \textbf{2) deterministic sampling: }the most likely sequence of states is sampled and remains unchanged in successive sampling trials. We use the forward variable of HSMM for deterministic sampling from the model. The forward variable $\alpha^{\tp{HSMM}}_{t,i} \triangleq P(z_t = i, \mb{\xi}_1 \ldots \mb{\xi}_t | \theta)$ requires marginalizing over the duration steps along with all possible state sequences. The probability of a datapoint $\mb{\xi}_t$ to be in state $i$ at time step $t$ given the partial observation sequence $\{\mb{\xi}_1,\ldots,\mb{\xi}_t\}$ is now specified as \cite{Yu10}
\begin{equation}
  \alpha^{\tp{HSMM}}_{t,i} = \sum_{s=1}^{\min(s^{\max},t-1)}\sum\limits_{j=1}^K \alpha^{\tp{HSMM}}_{t-s,j}\; a_{j,i} \; \mathcal{N}(s | \mu_i^{S}, \Sigma_i^{S}) \prod_{c = t - s + 1}^{t}
	\mathcal{N}\big(\mb{\xi}_c|\;\mb{\mu}_i,\mb{\Sigma}_i\big), \label{Eq: alpha_HSMM}
\end{equation} where the initialization is given by $\alpha_{1,i}^{\tp{HSMM}} = \Pi_i\;\mathcal{N}(1 | \mu_i^{S}, \Sigma_i^{S})\; \mathcal{N	}\big(\mb{\xi}_1|\;\mb{\mu}_i,\mb{\Sigma}_i\big)$, and the output distribution in state $i$ is conditionally independent for the $s$ duration steps given as $\prod_{c = t - s + 1}^{t} \mathcal{N}\big(\mb{\xi}_c|\;\mb{\mu}_i,\mb{\Sigma}_i\big)$. Note that for $t < s^{\max}$, the sum over duration steps is computed for $t-1$ steps, instead of $s^{\max}$. Without the observation sequence for the next time steps, the forward variable simplifies to
\begin{equation}
\alpha_{t,i}^{\tp{HSMM}} = \sum_{s=1}^{\min(s^{\max},t-1)}\sum\limits_{j=1}^K \alpha^{\tp{HSMM}}_{t-s,j}\; a_{j,i} \; \mathcal{N}(s | \mu_i^{S}, \Sigma_i^{S}).
\end{equation} The forward variable is used to plan the movement sequence for the next $T_p$ steps with $t = t+1 \ldots T_p$. During prediction, we only use the transition matrix and the duration model to plan the future evolution of the initial/current state and omit the influence of the spatial data that we cannot observe, i.e., $\mathcal{N}(\mb{\xi}_t|\mb{\mu}_i, \mb{\Sigma}_i) = 1$ for $t > 1$. This is used to retrieve a step-wise reference trajectory $\mathcal{N}(\mb{\hat{\mu}}_t, \mb{\hat{\Sigma}}_t)$ from a given state sequence $\mb{z}_t$ computed from the forward variable with,
\begin{equation}
\mb{z}_t = \{z_{t}, \ldots, z_{T_p} \} = \argmax_i \; \alpha^{\tp{HSMM}}_{t,i}, \quad \mb{\hat{\mu}}_t = \mb{\mu}_{z_t}, \quad \mb{\hat{\Sigma}}_t = \mb{\Sigma}_{z_t}.
\end{equation}
Fig.\ \ref{fig: HSMMLQT} shows a conceptual representation of the step-wise sequence of states generated by deterministically sampling from HSMM encoding of the Z-shaped data. In the next section, we show how to synthesise robot movement from this step-wise sequence of states in a smooth manner.
%
\subsection{Motion Generation with Linear Quadratic Tracking}
We formulate the motion generation problem given the step-wise desired sequence of states $\{\mathcal{N}(\mb{\hat{\mu}}_t, \mb{\hat{\Sigma}}_t)\}_{t=1}^{T_p}$ as sequential optimization of a scalar cost function with a linear quadratic tracker (LQT) \cite{borrelli11}. The control policy $\mb{u}_t$ at each time step is obtained by minimizing the cost function over the \textbf{finite time horizon} $T_p$,
\begin{gather}
c_t(\mb{\xi}_t, \mb{u}_t) = \sum_{t=1}^{T_p} (\mb{\xi}_t - \mb{\hat{\mu}}_t)^{\trsp} \mb{Q}_t (\mb{\xi}_t - \mb{\hat{\mu}}_t) + \mb{u}_t^{\trsp} \mb{R}_t \mb{u}_t, \label{Eq: CostLQT} \\
\mathrm{s.t.} \quad \mb{\xi}_{t+1} = \mb{A}_d \mb{\xi}_t + \mb{B}_d \mb{u}_t, \nonumber 
\end{gather} starting from the initial state $\mb{\xi}_1$ and following the discrete linear dynamical system specified by $\mb{A}_d$ and $\mb{B}_d$. We consider a linear time-invariant double integrator system to describe the system dynamics. Alternatively, a time-varying linearization of the system dynamics along the reference trajectory can also be used to model the system dynamics without loss of generality. Both discrete and continuous time linear quadratic regulator/tracker can be used to follow the desired trajectory. The discrete time formulation, however, gives numerically stable results for a wide range of values of $\mb{R}$. The control law $\mb{u}_t^{*}$ that minimizes the cost function in Eq. \eqref{Eq: CostLQT} under finite horizon subject to the linear dynamics in discrete time is given as,
\begin{equation}
\mb{u}_t^{*} = \mb{K}_t (\mb{\hat{\mu}}_t - \mb{\xi}_t) + \mb{u}_t^{\text{FF}},
\end{equation}
where $\mb{K}_t = [\mb{K}^{\ty{P}}_t, \mb{K}^{\ty{V}}_t]$ are the full stiffness and damping matrices for the feedback term, and $\mb{u}_t^{\text{FF}}$ is the feedforward term (see App. \ref{app: LQT_deriv} for computing the gains). Fig. \ref{fig: HSMMLQT} shows the results of applying discrete LQT on the desired step-wise sequence of states sampled from an HSMM encoding of the Z-shaped demonstrations. Note that the gains can be precomputed before simulating the system  if the reference trajectory does not change during the reproduction of the task. The resulting trajectory $\mb{\xi}_t^{*}$ smoothly tracks the step-wise reference trajectory $\mb{\hat{\mu}}_t$ and the gains $\mb{K}^{\ty{P}}_t, \mb{K}^{\ty{V}}_t$ locally stabilize $\mb{\xi}_t$ along $\mb{\xi}_t^{*}$ in accordance with the precision required during the task.
\begin{figure*}[tbp]
\centering
\includegraphics[trim={3.4cm 2.7cm 3cm 2.3cm},clip,scale=0.43]{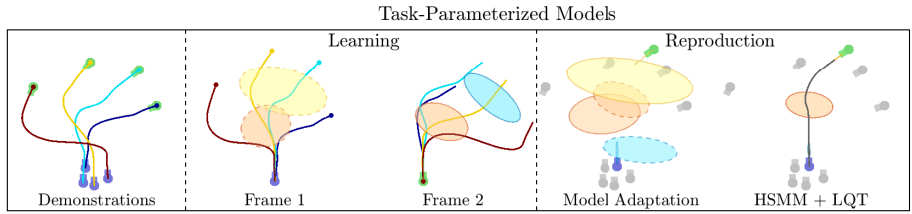}
\caption{\small Task-parameterized formulation of HSMM: four demonstrations on left are observed from two coordinate systems that define the start and end position of the demonstration (starting in purple position and ending in green position for each demonstration). The generative model is learned in the respective coordinate systems. The model parameters in respective coordinate systems are adapted to the new unseen object positions by computing the products of linearly transformed Gaussian mixture components. The resulting HSMM is combined with LQT for smooth retrieval of manipulation tasks.}\label{fig: taskParam}
\end{figure*}
\section{Invariant Task-Parameterized HSMMs} \label{sec: TPHMM}
Conventional approaches to encode task variations such as change in the pose of the object is to augment the state of the environment with the policy parameters \cite{Paraschos17}. Such an encoding, however, does not capture the geometric structure of the problem. Our approach exploits the problem structure by introducing the task parameters in the form of coordinate systems that observe the demonstrations from multiple perspectives. Task-parameterization enables the model parameters to adapt in accordance with the external task parameters that describe the environmental situation, instead of hard coding the solution for each new situation or handling it in an \textit{ad hoc} manner \cite{Wilson99}. When a different situation occurs (pose of the object changes), changes in the task parameters/reference frames are used to modulate the model parameters in order to adapt the robot movement to the new situation. 

\subsection{Model Learning}

We represent the task parameters with $F$ coordinate systems, defined by $\{\mb{A}_{j}, \mb{b}_{j}\}_{j=1}^{F}$, where $\mb{A}_{j}$ denotes the orientation of the frame as a rotation matrix and $\mb{b}_{j}$ represents the origin of the frame. We assume that the coordinate frames are specified by the user, based on prior knowledge about the carried out task. Typically, coordinate frames will be attached to objects, tools or locations that could be relevant in the execution of a task. Each datapoint $\mb{\xi}_t$ is observed from the viewpoint of $F$ different experts/frames, with $\mb{\xi}_t^{(j)} = \mb{A}_{j}^{-1} (\mb{\xi}_t - \mb{b}_{j})$ denoting the datapoint observed with respect to frame $j$. The parameters of the task-parameterized HSMM are defined by \[\theta = \Big \{ \{\mb{\mu}_{i}^{(j)},\mb{\Sigma}_{i}^{(j)}\}_{j=1}^F, \{a_{i,m} \}_{m=1}^{K}, \mu_{i}^{S},\Sigma_{i}^{S} \Big\}_{i=1}^{K},\] where $\mb{\mu}_{i}^{(j)}$ and $\mb{\Sigma}_{i}^{(j)}$ define the mean and the covariance matrix of $i$-th mixture component in frame $j$. Parameter updates of the task-parameterized HSMM algorithm remain the same as HSMM, except the computation of the mean and the covariance matrix is repeated for each coordinate system separately. The emission distribution of the $i$-th state is represented by the product of the probabilities of the datapoint to belong to the $i$-th Gaussian in the corresponding $j$-th coordinate system. The forward variable of HMM in the task-parameterized formulation is described as
\begin{equation}
  \alpha^{\tp{TP-HMM}}_{t,i} = \Big(\sum\limits_{j=1}^K \alpha^{\tp{HMM}}_{t-1,j}\; a_{j,i}\Big) \;
	\prod_{j=1}^{F} \mathcal{N}\big(\mb{\xi}_t^{(j)}|\;\mb{\mu}_i^{(j)},\mb{\Sigma}_i^{(j)}\big). \label{Eq: alpha_TP_HMM}
\end{equation} Similarly, the backward variable $\beta_{t,i}^{\tp{TP-HMM}}$, the smoothed node marginal $\gamma_{t,i}^{\tp{TP-HMM}}$, and the smoothed edge marginal $\zeta_{t,i,j}^{\tp{TP-HMM}}$ can be computed by replacing the emission distribution $\mathcal{N}(\mb{\xi}_t |\; \mb{\mu}_i, \mb{\Sigma}_i)$ with the product of probabilities of the datapoint in each frame $\prod_{j=1}^{F} \mathcal{N}\big(\mb{\xi}_t^{(j)}|\;\mb{\mu}_i^{(j)},\mb{\Sigma}_i^{(j)}\big)$. The duration model $\mathcal{N}(s|\mu_i^{S},\Sigma_i^{S})$ is used as a replacement of the self-transition probabilities $a_{i,i}$. The hidden state sequence over all demonstrations is used to define the duration model parameters $\{\mu_i^{S}, \Sigma_i^{S}\}$ as the mean and the standard deviation of staying $s$ consecutive time steps in the $i$-th state.

\subsection{Model Adaptation in New Situations}
In order to combine the output from coordinate frames of reference for an unseen situation represented by the frames $\{\mb{\tilde{A}}_{j}, \mb{\tilde{b}}_{j} \}_{j=1}^F$, we linearly transform the Gaussians back to the global coordinates with $\{\mb{\tilde{A}}_{j}, \mb{\tilde{b}}_{j} \}_{j=1}^F$, and retrieve the new model parameters $\{\mb{\tilde{\mu}}_{i}, \mb{\tilde{\Sigma}}_{i}\}$ for the $i$-th mixture component by computing the products of the linearly transformed Gaussians (see Fig. \ref{fig: taskParam}) 
\begin{equation}
	\label{Eq: ProdGaussEq} 
	\mathcal{N}(\mb{\tilde{\mu}}_{i}, \mb{\tilde{\Sigma}}_{i}) \;\propto\;
  \prod\limits_{j=1}^F \mathcal{N}\left(\mb{\tilde{A}}_{j} \mb{\mu}_{i}^{(j)} + 
  \mb{\tilde{b}}_{j}, \mb{\tilde{A}}_{j} \mb{\Sigma}_{i}^{(j)} \mb{\tilde{A}}_{j}^{\trsp} \right). 
\end{equation} 

Evaluating the products of Gaussians represents the observation distribution of HSMM, whose output sequence is decoded and combined with LQT for smooth motion generation as shown in the previous section.
\begin{equation}
\mb{\tilde{\Sigma}}_i  =  \left( \sum_{j=1}^{F} \left(\mb{\tilde{A}}_j \mb{\Sigma}_i^{(j)} \mb{\tilde{A}}_j^{\trsp}\right)^{-1} \right)^{-1}, \qquad 
\mb{\tilde{\mu}}_i = \mb{\tilde{\Sigma}}_i \sum_{j=1}^{F}\left(\mb{\tilde{A}}_j \mb{\Sigma}_i^{(j)} \mb{\tilde{A}}_j^{\trsp}\right)^{-1} \left(\mb{\tilde{A}}_j \mb{\mu}_i^{(j)} +  \mb{\tilde{b}}_j \right). \label{Eq: GeneralizedSigma}
\end{equation}
\section{Latent Space Representations} \label{sec: LS-HMM}
Dimensionality reduction has long been recognized as a fundamental problem in unsupervised learning. Model-based generative models such as HSMMs tend to suffer from the \textit{curse of dimensionality} when few datapoints are available. We use statistical subspace clustering methods that reduce the number of parameters to be robustly estimated to address this problem. A simple way to reduce the number of parameters would be to constrain the covariance structure to a diagonal or spherical/isotropic matrix, and restrict the number of parameters at the cost of treating each dimension separately. Such decoupling, however, cannot encode the important motor control principles of coordination, synergies and action-perception couplings \cite{Wolpert11}.

Consequently, we seek out a latent feature space in the high-dimensional data to reduce the number of model parameters that can be robustly estimated. We consider three formulations to this end: 1) low-rank decomposition of the covariance matrix using \textit{Mixture of Factor Analyzers} (MFA) approach \cite{McLachlan03}, 2) partial tying of the covariance matrices of the mixture model with the same set of basis vectors, albeit different scale with semi-tied covariance matrices \cite{Gales99,Tanwani16}, and 3) Bayesian non-parametric sequence clustering under small variance asymptotics \cite{Kulis12,Roychowdhury13,Tanwani16a}. All the decompositions can readily be combined with invariant task-parameterized HSMM and LQT for encapsulating reactive autonomous behaviour as shown in the previous section.
\subsection{Mixture of Factor Analyzers}
\begin{figure}[tbp]
\centering
\includegraphics[trim={0.3cm 7.0cm 0.3cm 7.0cm},clip,scale = 0.36]{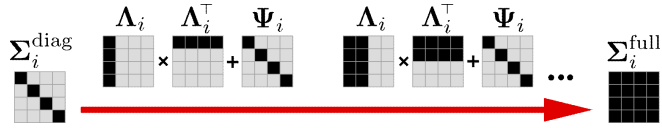}
\caption{\small Parameters representation of a diagonal, full and mixture of factor analyzers decomposition of covariance matrix. Filled blocks represent non-zero entries.} \label{fig: mfa_represent}
\end{figure}

The basic idea of MFA is to perform subspace clustering by assuming the covariance structure for each component of the form,
\begin{equation}
  \mb{\Sigma}_i = \mb{\Lambda}_i\mb{\Lambda}_i^\trsp+\mb{\Psi}_i ,
\end{equation}
where $\mb{\mb{\Lambda}_i}\in\mathbb{R}^{D\times d}$ is the \emph{factor loadings matrix} with $d\!<\!D$ for parsimonious representation of the data, and $\mb{\Psi}_i$ is the diagonal noise matrix (see Fig.\ \ref{fig: mfa_represent} for MFA representation in comparison to a diagonal and a full covariance matrix). Note that the mixture of probabilistic principal component analysis (MPPCA) model is a special case of MFA with the distribution of the errors assumed to be isotropic with $\mb{\Psi}_i\!=\!\mb{I}\sigma_i^2$ \cite{Tipping99}.
The MFA model assumes that $\mb{\xi}_t$ is generated using a linear transformation of $d$-dimensional vector of latent (unobserved) factors $\mb{f}_t$,
\begin{equation}
  \mb{\xi}_t = \mb{\Lambda}_i \mb{f}_t + \mb{\mu}_i + \mb{\epsilon},
\end{equation}
where $\mb{\mu}_i\in\mathbb{R}^D$ is the mean vector of the $i$-th factor analyzer, $\mb{f}_t\!\sim\!\mathcal{N}(\mb{0},\mb{I})$  is a normally distributed factor, and $\mb{\epsilon}\!\sim\!\mathcal{N}(\mb{0},\mb{\Psi}_i)$ is a zero-mean Gaussian noise with diagonal covariance $\mb{\Psi}_i$. The diagonal assumption implies that the observed variables are independent given the factors.   
Note that the subspace of each cluster is not spanned by orthogonal vectors, whereas it is a necessary condition in models based on eigendecomposition such as PCA. Each covariance matrix of the mixture component has its own subspace spanned by the basis vectors of $\mb{\Sigma}_i$. As the number of components increases to encode more complex skills, an increasing large number of potentially redundant parameters are used to fit the data. Consequently, there is a need to share the basis vectors across the mixture components as shown below by semi-tying the covariance matrices of the mixture model.
%
\subsection{Semi-Tied Mixture Model}
When the covariance matrices of the mixture model share the same set of parameters for the latent feature space, we call the model a \textit{semi-tied} mixture model \cite{Tanwani16}. The main idea behind semi-tied mixture models is to decompose the covariance matrix $\mb{\Sigma}_i$ into two terms: a common latent feature matrix $\mb{H} \in \mathbb{R}^{D\times D}$ and a component-specific diagonal matrix $\mb{\Sigma}_i^{(\diag)} \in\mathbb{R}^{D\times D}$, i.e.,
\begin{equation}\label{eq: SigmaGMM}
\mb{\Sigma}_i = \mb{H} \mb{\Sigma}_i^{(\diag)} \mb{H}^\trsp.
\end{equation} 
The latent feature matrix encodes the locally important synergistic directions represented by $D$ non-orthogonal basis vectors that are shared across all the mixture components, while the diagonal matrix selects the appropriate subspace of each mixture component as convex combination of a subset of the basis vectors of $\mb{H}$. Note that the eigen decomposition of $\mb{\Sigma}_i = \mb{U}_{i} \mb{\Sigma}_i^{(\diag)} \mb{U}_{i}^{\trsp} $ contains $D$ basis vectors of $\mb{\Sigma}_i$ in $\mb{U}_i$. In comparison, semi-tied mixture model gives $D$ globally representative basis vectors that are shared across all the mixture components. The parameters $\mb{H}$ and $\mb{\Sigma}_i^{(\diag)}$ are updated in closed form with EM updates of HSMM \cite{Gales99}.

The underlying hypothesis in semi-tying the model parameters is that similar coordination patterns occur at different phases in a manipulation task. By exploiting the spatial and temporal correlation in the demonstrations, we reduce the number of parameters to be estimated while locking the most important synergies to cope with perturbations. This allows the reuse of the discovered synergies in different parts of the task having similar coordination patterns. In contrast, the MFA decomposition of each covariance matrix separately cannot exploit the temporal synergies, and has more flexibility in locally encoding the data.
\subsection{Bayesian Non-Parametrics under Small Variance Asymptotics} \label{sec: BNP-HMM}
Specifying the number of latent states in a mixture model is often difficult. Model selection methods such as cross-validation or Bayesian Information Criterion (BIC) are typically used to determine the number of states. Bayesian non-parametric approaches comprising of Hierarchical Dirichlet Processes (HDPs) provide a principled model selection procedure by Bayesian inference in an HMM with infinite number of states \cite{Teh06}. These approaches provide flexibility in model selection, however, their widespread use is limited by the computational overhead of existing sampling-based and variational techniques for inference. We take a \textbf{small variance asymptotics} approximation of the Bayesian non-parametric model that collapses the posterior to a simple deterministic model, while retaining the non-parametric characteristics of the algorithm.
\begin{figure*}[tbp]
\centering
\includegraphics[trim={1.6cm 3.7cm 2.0cm 4.2cm},clip,scale = 0.47]{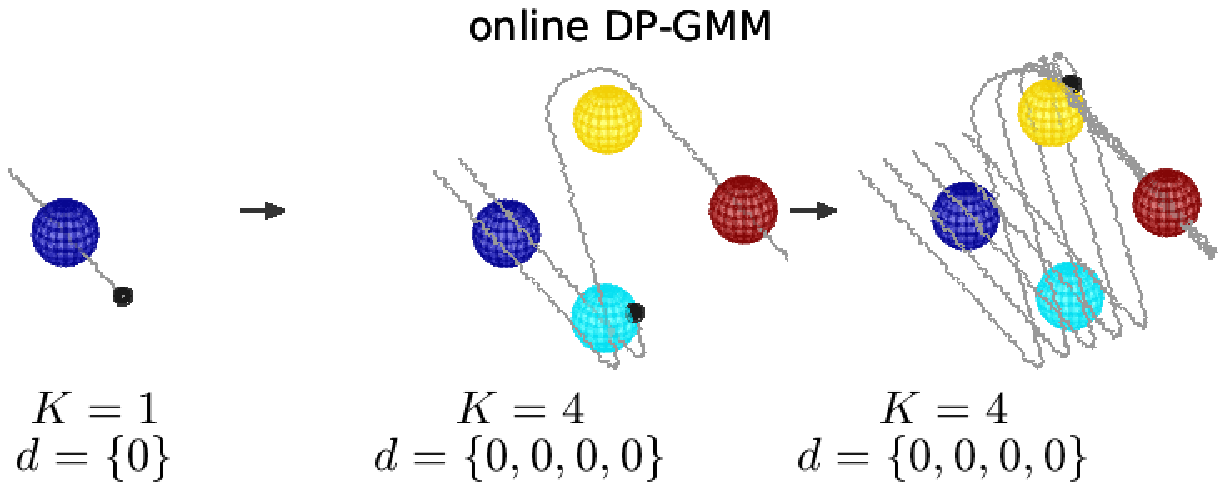}\rulesep 
\includegraphics[trim={1.6cm 3.7cm 2.0cm 4.2cm},clip,scale = 0.47]{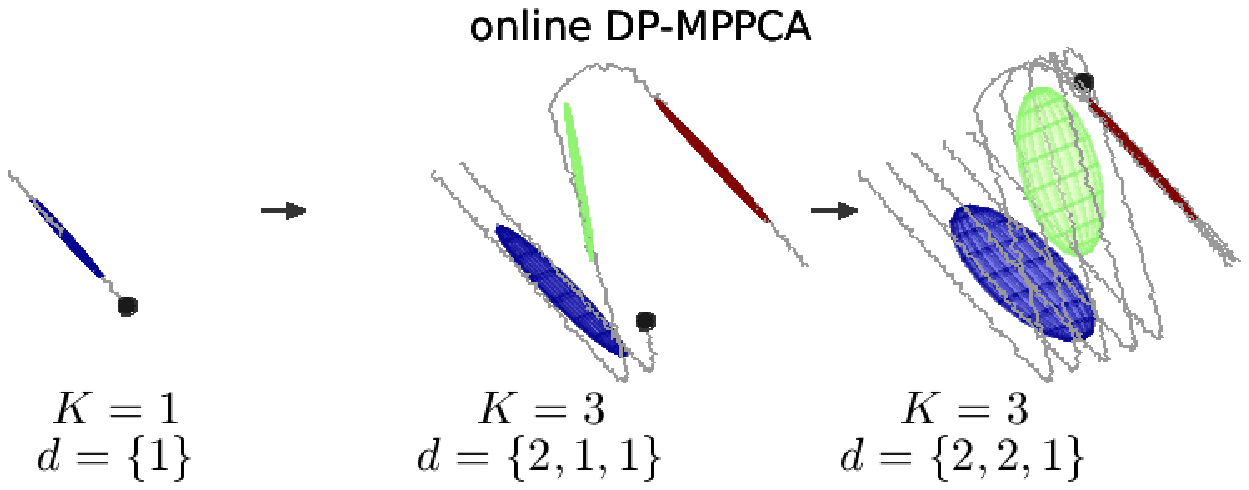}
\caption{\small Bayesian non-parametric clustering of Z-shaped streaming data under small variance asymptotics with: \textit{(left)} online DP-GMM, \textit{(right)} online DP-MPPCA. Note that the number of clusters and the subspace dimension of each cluster is adapted in a non-parametric manner.}\label{fig: Z_Coceptual}
\end{figure*}

Small variance asymptotic (SVA) analysis implies that the covariance matrix $\mb{\Sigma}_{i}$ of all the Gaussians is set to the isotropic noise $\sigma^{2}$, i.e., $\mb{\Sigma}_{i} \approx \lim_{\sigma^{2} \rightarrow 0} \sigma^{2} \mb{I}$ in the likelihood function and the prior distribution \cite{Kulis12,Broderick13}. The analysis yields simple deterministic models, while retaining the non-parametric nature. For example, SVA analysis of the Bayesian non-parametric GMM leads to the DP-means algorithm \cite{Kulis12}. Similarly, SVA analysis of the Bayesian non-parametric HMM under Hierarchical Dirichlet Process (HDP) yields the segmental $k$-means problem \cite{Roychowdhury13}.

Restricting the covariance matrix to an isotropic/spherical noise, however, fails to encode the coordination patterns in the demonstrations. Consequently, we model the covariance matrix in its intrinsic affine subspace of dimension $d_{i}$ with projection matrix $\mb{\Lambda}_{i}^{d_{i}} \in \mathbb{R}^{D \times d_{i} }$, such that $d_{i} < D$ and $\mb{\Sigma}_{i} = \lim_{\sigma^{2} \rightarrow 0} \mb{\Lambda}_{i}^{d_{i}} \mb{\Lambda}_{i}^{{d_{i}}^{\trsp}} + \sigma^{2} \mb{I}$ (akin to DP-MPPCA model). Under this assumption, we apply the small variance asymptotic limit on the remaining $(D - d_{i})$ dimensions to encode the most important coordination patterns while being parsimonious in the number of parameters (see Fig. \ref{fig: Z_Coceptual}). Performing small-variance asymptotics of the joint likelihood of HDP-HMM yields the maximum aposteriori estimates of the parameters by iteratively minimizing the loss function\footnote{Setting $d_i = 0$ by choosing $\lambda_1 \gg 0$ gives the loss function formulation with isotropic Gaussian under small variance asymptotics \cite{Roychowdhury13}.}
\begin{multline*}
	\mathcal{L} (\mb{z}, \mb{d}, \mb{\mu}, \mb{U}, \mb{a}) = 
	\sum_{t=1}^{T} \text{dist}(\mb{\xi}_t, \mb{\mu}_{z_t} , \mb{U}_{z_t}^{d_i})^{2} + \lambda(K - 1 ) \\ 
	+ \lambda_1 \sum_{i=1}^{K}d_i - \lambda_2 \sum_{t=1}^{T-1} 
	\log(a_{z_{t}, z_{t+1}}) + \lambda_3 \sum_{i=1}^{K} (\tau_{i} - 1),
\end{multline*} 
where $\text{dist}(\mb{\xi}_t, \mb{\mu}_{z_t} , \mb{U}_{z_t}^{d})^{2}$ represents the distance of the datapoint $\mb{\xi}_t$ to the subspace of cluster $z_t$ defined by mean $\mb{\mu}_{z_t}$ and unit eigenvectors of the covariance matrix $\mb{U}_{z_t}^{d}$ (see App. \ref{app: dist_subspace}). The algorithm optimizes the number of clusters and the subspace dimension of each cluster while minimizing the distance of the datapoints to the respective subspaces of each cluster. The $\lambda_2$ term favours the transitions to states with higher transition probability (states which have been visited more often before), $\lambda_3$ penalizes for transition to unvisited states with $\tau_{i}$ denoting the number of distinct transitions out of state $i$, while $\lambda$ and $\lambda_1$ are the penalty terms for increasing the number of states and the subspace dimension of each output state distribution. 

The analysis is used here for scalable online sequence clustering that is non-parametric in the number of clusters and the subspace dimension of each cluster. The resulting algorithm groups the data in its low dimensional subspace with non-parametric mixture of probabilistic principal component analyzers based on Dirichlet process, and captures the state transition and state duration information in a HDP-HSMM. The cluster assignment and the parameter updates at each iteration minimize the loss function, thereby, increasing the model fitness while penalizing for new transitions, new dimensions and/or new clusters. An interested reader can find more details of the algorithm in \cite{Tanwani16a}.
\begin{figure}[tbp]
\centering
\includegraphics[trim={2.7cm 3.0cm 2.5cm 3.0cm},clip,scale = 0.3]{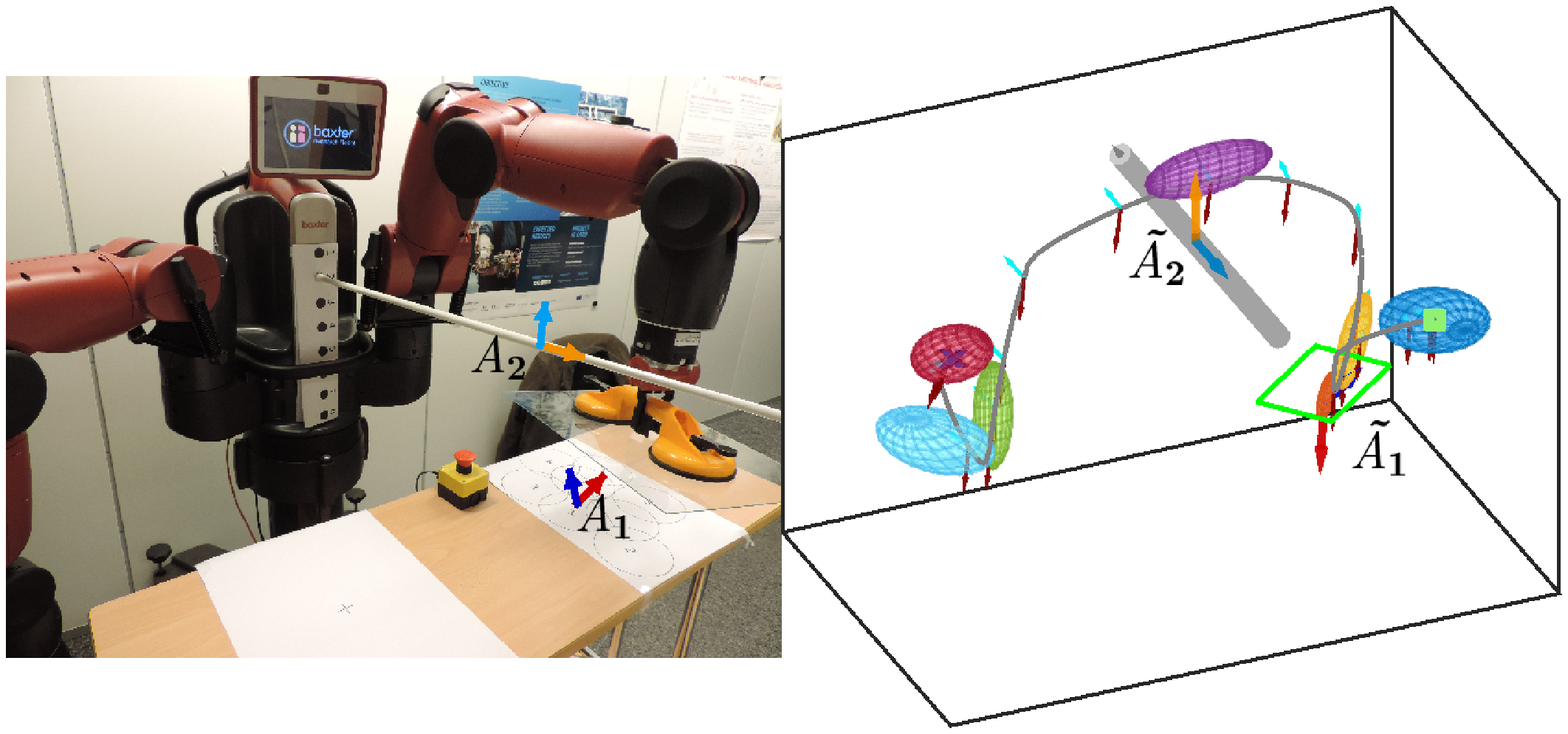}
\includegraphics[trim={2.0cm 0.78cm 1.4cm 1.5cm},clip,scale = 0.38]{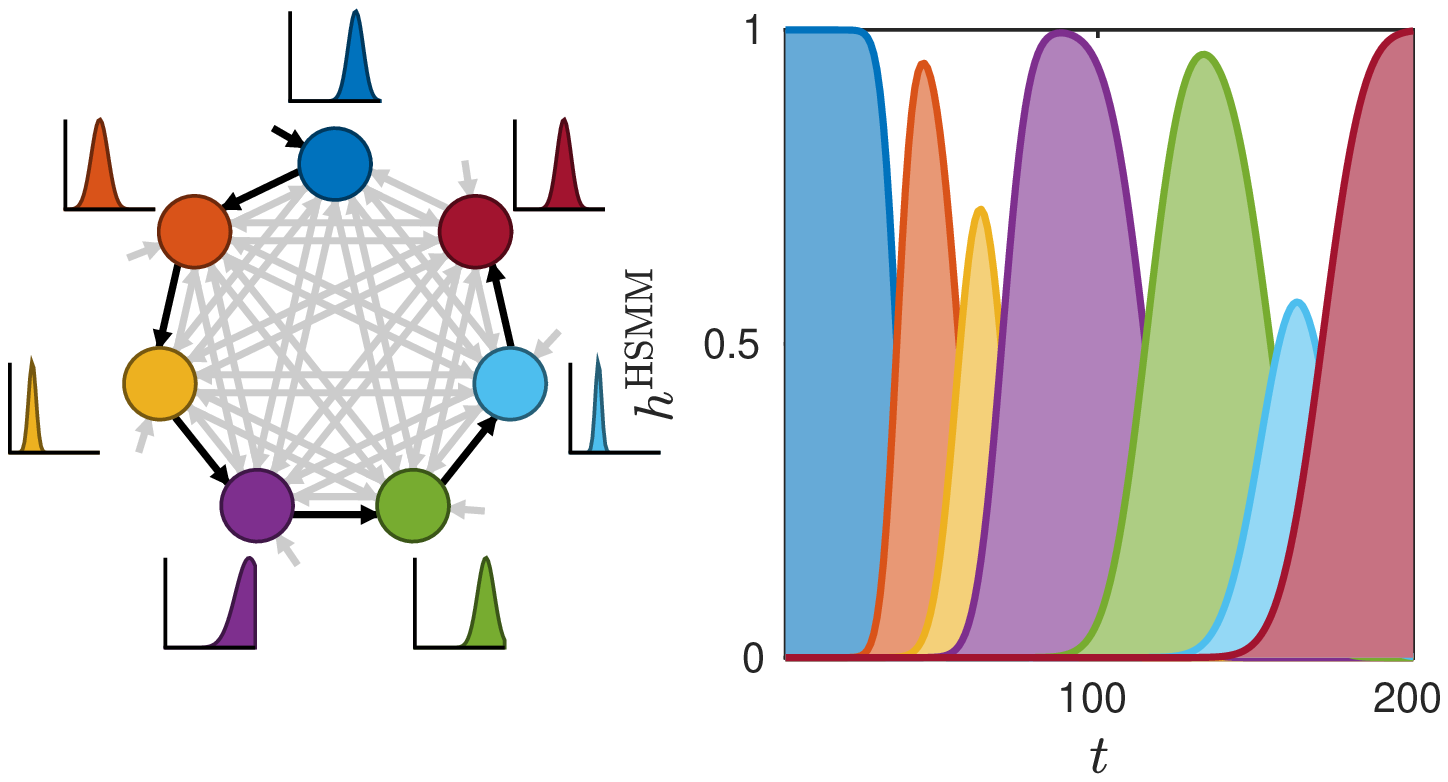}
\caption{\small \textit{(left)} Baxter robot picks the glass plate with a suction lever and places it on the cross after avoiding an obstacle of varying height, \textit{(centre-left)} reproduction for previously unseen object and obstacle position, \textit{(cente-right)} left-right HSMM encoding of the task with duration model shown next to each state ($s^{\max} = 100$), \textit{(right)} rescaled forward variable evolution of the forward variable over time.} \label{fig: Baxter_PP}
\end{figure}
\begin{figure*}[tbp]
\centering
\includegraphics[trim={4.8cm 3.0cm 4.8cm 2.8cm},clip,scale = 0.29]{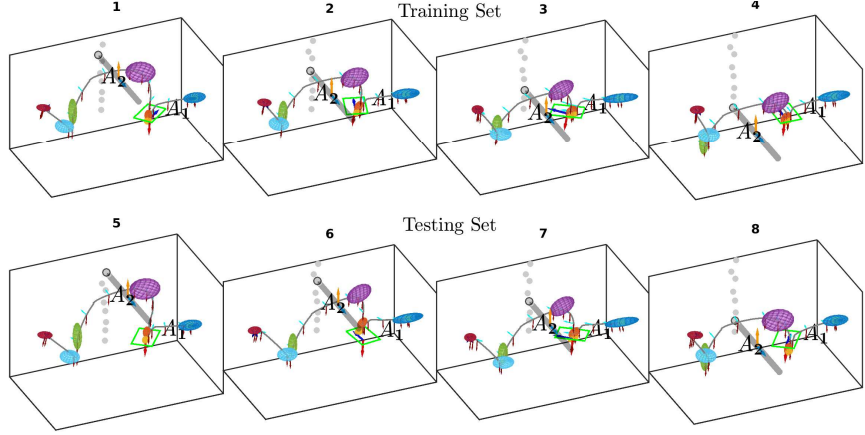} 
\caption{\small Task-Parameterized HSMM performance on pick-and-place with obstacle avoidance task: \textit{(top)} training set reproductions, \textit{(bottom)} testing set reproductions.} \label{fig: Baxter_PP_results}
\end{figure*}

\section{Experiments, Results and Discussion} \label{sec: Exp}
We now show how our proposed work enables a Baxter robot to learn a pick-and-place task from a few human demonstrations. The objective of the task is to place the object in a desired target position by picking it from different initial poses of the object, while adapting the movement to avoid the obstacle. The setup of pick-and-place task with obstacle avoidance is shown in Fig. \ref{fig: Baxter_PP}. The Baxter robot is required to grasp the glass plate with a suction lever placed in an initial configuration as marked on the setup. The obstacle can be vertically displaced to one of the  $8$ target configurations. We describe the task with two frames, one frame for the object initial configuration with $\{\mb{A}_1, \mb{b}_1\}$ and other frame for the obstacle $\{\mb{A}_2, \mb{b}_2\}$ with $\mb{A}_2 = \mb{I}$ and $\mb{b}_2$ to specify the centre of the obstacle. We collect $8$ kinesthetic demonstrations with different initial configurations of the object and the obstacle successively displaced upwards as marked with the visual tags in the figure. Alternate demonstrations are used for the training set, while the rest are used for the test set. Each observation comprises of the end-effector Cartesian position, quaternion orientation, gripper status (open/closed), linear velocity, quaternion derivative, and gripper status derivative with $D = 16, P=2$, and a total of $200$ datapoints per demonstration. We represent the frame $\{\mb{A}_1, \mb{b}_1\}$ as 
\begin{equation} \label{eq: FramesTP}
\mb{A}_1^{(n)} = \begin{bmatrix}
\mb{R}_1^{(n)} 		& 	\mb{0}	& 	\mb{0}	& \mb{0}	 & 0 \\ 
\mb{0}	&  	\mathbf{\mathcal{E}}_1^{(n)} 	& \mb{0}	 & \mb{0} & 0\\
\mb{0}	&	\mb{0} & \mb{R}_1^{(n)} 	& \mb{0} & 0 \\
\mb{0} 	&  	\mb{0} & 	\mb{0}	& \mathbf{\mathcal{E}}_1{(n)} & 0 \\
0 	&  	0 & 	0	& 0 & 1
 		\end{bmatrix},
  \mb{b}_1^{(n)} = \begin{bmatrix}
    \mb{p}_1^{(n)}\\
\mb{0} \\
\mb{0} \\
\mb{0} \\
1 
  \end{bmatrix}  
  , 		
\end{equation} where $\mb{p}_1^{(n)} \in \mathbb{R}^3, \mb{R}_1^{(n)} \in \mathbb{R}^{3 \times 3}, \mathbf{\mathcal{E}}_1^{(n)} \in \mathbb{R}^{4 \times 4}$ denote the Cartesian position, the rotation matrix and the quaternion matrix in the $n$-th demonstration respectively. Note that we do not consider time as an explicit variable as the duration model in HSMM encapsulates the timing information locally.

Performance setting in our experiments is as follows: $\{\pi_i,\mb{\mu}_i,\mb{\Sigma}_i\}_{i=1}^K$ are initialized using k-means clustering algorithm, $\mb{R} = 9 \mb{I}$,  where $\mb{I}$ is the identity matrix, learning converges when the difference of log-likelihood between successive demonstrations is less than $1\times10^{-4}$. Results of regenerating the movements with $7$ mixture components are shown in Fig. \ref{fig: Baxter_PP_results}. For a given initial configuration of the object, the model parameters are adapted by evaluating the product of Gaussians for a new frame configuration. The reference trajectory is then computed from the initial position of the robot arm using the forward variable of HSMM and tracked using LQT. The robot arm moves from its initial configuration to align itself with the first frame $\{\mb{A}_1, \mb{b}_1\}$ to grasp the object, and follows it with the movement to avoid the obstacle and subsequently, align with the second frame $\{\mb{A}_2, \mb{b}_2\}$ before placing the object and returning to a neutral position. The model exploits variability in the observed demonstrations to statistically encode different phases of the task such as reach, grasp, move, place, return. The imposed structure with task-parameters and HSMM allows us to acquire a new task in a few human demonstrations, and  generalize effectively in picking and placing the object. 
\begin{figure*}[tbp]
\centering
\includegraphics[trim={4.3cm 0.70cm 4.3cm 0.1cm},clip,scale = 0.333]{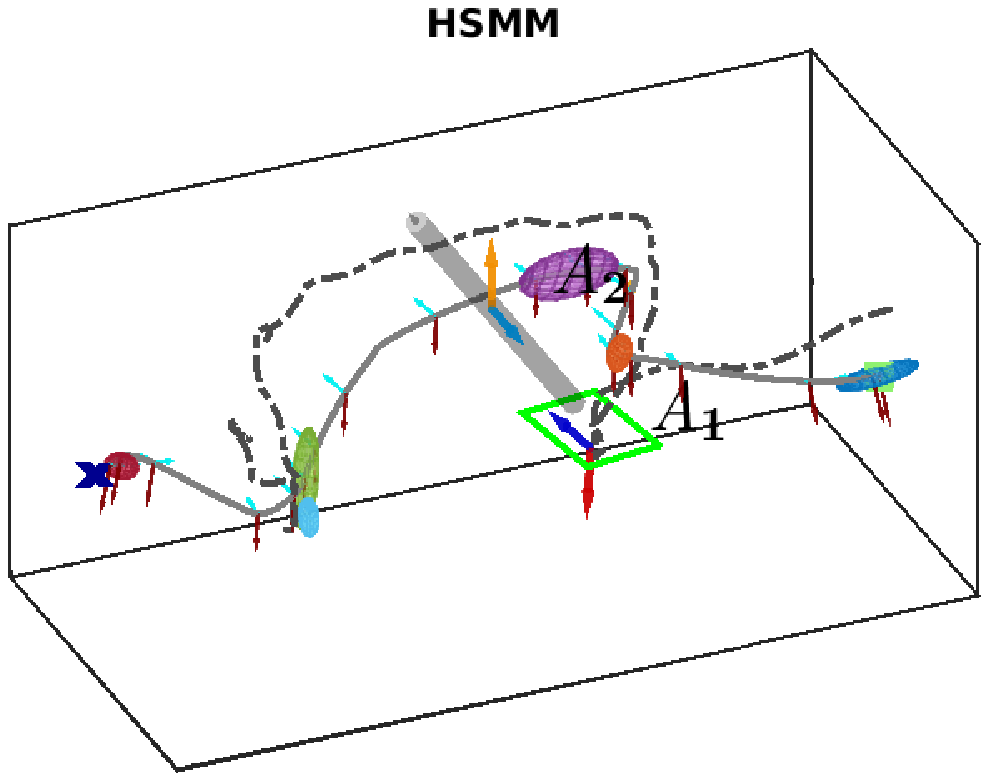} 
\includegraphics[trim={4.9cm 0.70cm 4.9cm 0.1cm},clip,scale = 0.333]{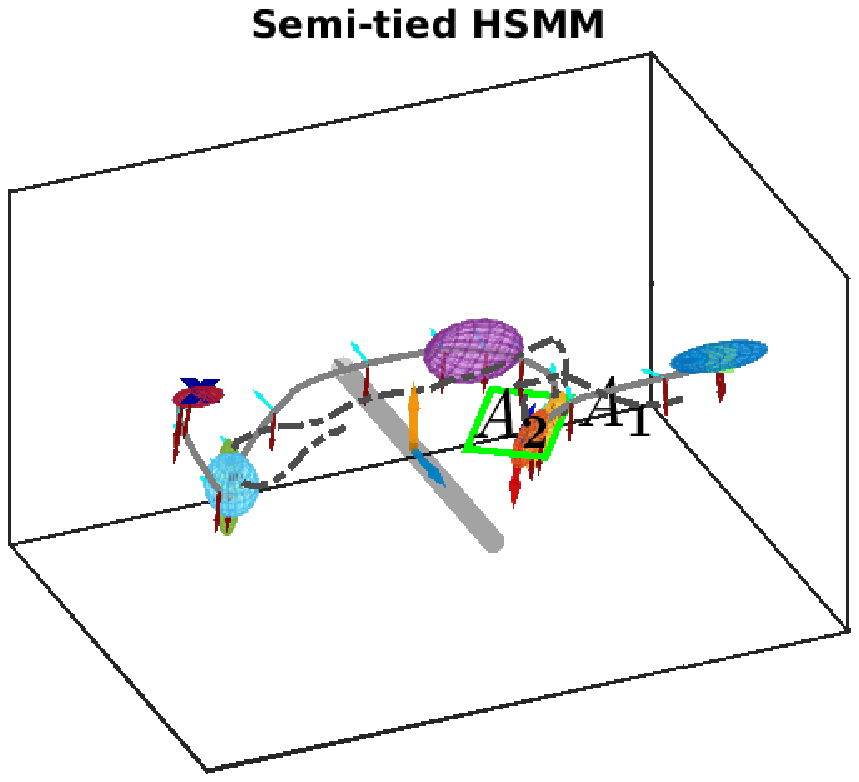}
\includegraphics[trim={4.9cm 0.70cm 4.9cm 0.1cm},clip,scale = 0.333]{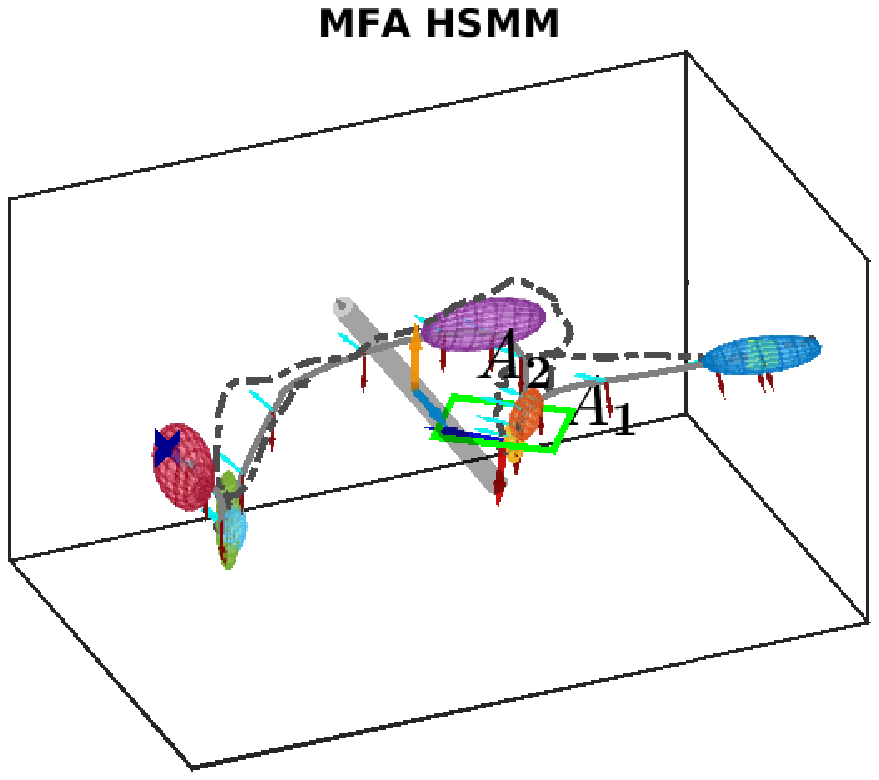}
\includegraphics[trim={5.1cm 0.70cm 5.0cm 0.1cm},clip,scale = 0.333]{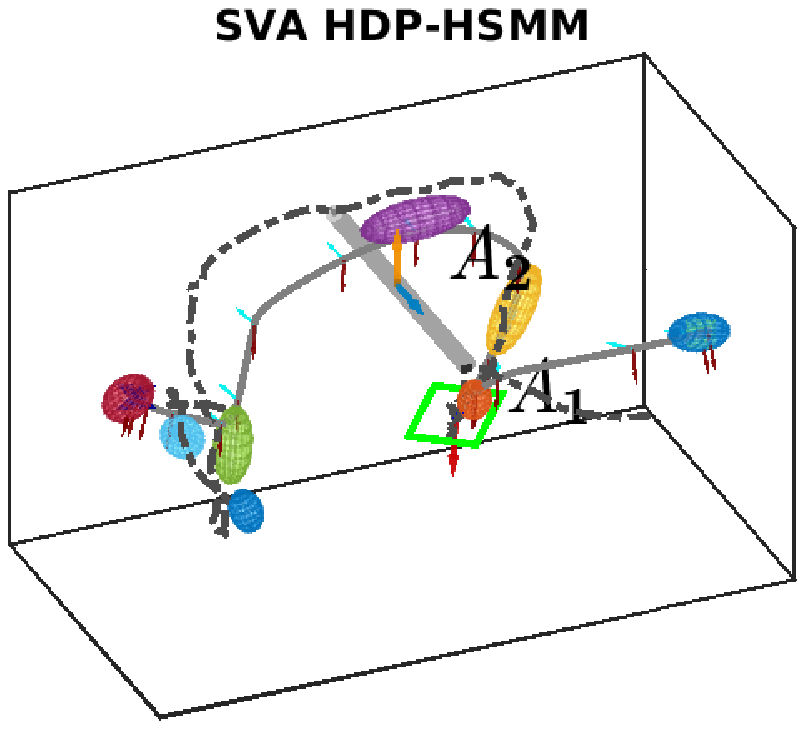}
\caption{\small Latent space representations of invariant task-parameterized HSMM for a randomly chosen demonstration from the test set. Black dotted lines show human demonstration, while grey line shows the reproduction from the model (see supplementary materials for details).} \label{fig: Baxter_PP_results_LS}
\end{figure*}
\begin{table}[!tb]
\caption{\small Performance analysis of invariant hidden Markov models with training MSE, testing MSE, number of parameters for pick-and-place task. MSE (in meters) is computed between the demonstrated trajectories and the generated trajectories (lower is better). Latent space formulations give comparable task performance with much fewer parameters.} \normalsize \centering \label{tab_TMM}
\begin{tabular}{|c||c|c|c|}
\hline
\multirow{2}{*}{\textbf{Model}} & \multirow{2}{*}{\textbf{Training MSE}} & \multirow{2}{*}{\textbf{Testing MSE}} & \textbf{Number of} \\
&  &  & \textbf{Parameters}  \\\hline \hline
\multicolumn{4}{|c|}{\textbf{pick-and-place via obstacle avoidance $(K = 7, F = 2, D = 16)$}}\\\hline
HSMM & $\mathbf{0.0026 \pm 0.0009}$ & $0.014 \pm 0.0085$ & $2198$ \\\hline
Semi-Tied HSMM & $0.0033 \pm 0.0016$ & $0.0131 \pm 0.0077$ & $1030$ \\\hline
MFA HSMM ($d_k = 1$) & $0.0037 \pm 0.0011$ & $\mathbf{0.0109 \pm 0.0068}$ & $\mathbf{742}$ \\\hline 
MFA HSMM ($d_k = 4$) & $0.0025 \pm 0.0007$ & $0.0119 \pm 0.0077$ & $1414$ \\\hline
MFA HSMM ($d_k = 7$) & $0.0023 \pm 0.0009$ & $0.0123 \pm 0.0084$ & $2086$ \\\hline 
SVA HDP HSMM & \multirow{2}{*}{$0.0073 \pm 0.0024$} & \multirow{2}{*}{$0.0149 \pm 0.0072$} & \multirow{2}{*}{$1352$} \\
($K=8, \bar{d}_k = 3.94$) & & & \\\hline 
\end{tabular}
\end{table}	
Table \ref{tab_TMM} evaluates the performance of the invariant task-parameterized HSMM with latent space representations. We observe significant reduction in the model parameters, while achieving better generalization on the unseen situations compared to the task-parameterized HSMM with full covariance matrices (see Fig. \ref{fig: Baxter_PP_results_LS} for comparison across models). It is seen that the MFA decomposition gives the best performance on test set with much fewer parameters.
\section{Conclusions}
Learning from demonstrations is a promising approach to teach manipulation skills to robots. In contrast to deep learning approaches that require extensive training data, generative mixture models are useful for learning from a few examples that are not explicitly labelled. The formulations are inspired by the need to make generative mixture models easy to use for robot learning in a variety of applications, while requiring considerably less learning time. 

We have presented formulations for learning invariant task representations with hidden semi-Markov models for recognition, prediction, and reproduction of manipulation tasks; along with learning in latent space representations for robust parameter estimation of mixture models with high-dimensional data. By sampling the sequence of states from the model and following them with a linear quadratic tracking controller, we are able to autonomously perform manipulation tasks in a smooth manner. This has enabled a Baxter robot to tackle a pick-and-place via obstacle avoidance problem from previously unseen configurations of the environment. A relevant direction of future work is to not rely on specifying the task parameters manually, but to infer generalized task representations from the videos of the demonstrations in learning the invariant segments. Moreover, learning the task model from a  a small set of labelled demonstrations in a semi-supervised manner is an important aspect in extracting meaningful segments from demonstrations.
\newline\newline
\small{\textbf{{Acknowledgements}}: This work was, in large part, carried out at Idiap Research Institute and Ecole Polytechnique Federale de Lausanne (EPFL) Switzerland. This work was in part supported by the DexROV project through the EC Horizon 2020 program (Grant $635491$), and the NSF National Robotics Initiative Award $1734633$ on Scalable Collaborative Human-Robot Learning (SCHooL). The information, data, comments, and views detailed herein may not necessarily reflect the endorsements of the sponsors.}
\bibliographystyle{plain}
\bibliography{bibliography}
\clearpage
\normalsize
\section{Appendix}
\subsection{EM updates of HMM} \label{app: EM-HMM}
The intermediary variables, namely \textbf{forward variable} $\alpha^{\tp{HMM}}_{t,i}$, \textbf{backward variable} $\beta^{\tp{HMM}}_{t,i}$, \textbf{smoothed node marginal} $\gamma^{\tp{HMM}}_{t,i}$, and \textbf{smoothed edge marginal} $\zeta^{\tp{HMM}}_{t,i,j}$ are mathematically represented as:
\begin{align}
\alpha^{\tp{HMM}}_{t,i} &= \Big(\sum\limits_{j=1}^K \alpha^{\tp{HMM}}_{t-1,j}\; a_{j,i}\Big) \;
	\mathcal{N}\big(\mb{\xi}_t|\;\mb{\mu}_i,\mb{\Sigma}_i\big), & \beta^{\tp{HMM}}_{t,i} &= \sum\limits_{j=1}^K a_{i,j} \; \mathcal{N}\big(\mb{\xi}_{t+1}|\;\mb{\mu}_j,\mb{\Sigma}_j\big) \; \beta^{\tp{HMM}}_{t+1,j}, \nonumber \\
\gamma^{\tp{HMM}}_{t,i}
  &= \frac{\alpha^{\tp{HMM}}_{t,i}\beta^{\tp{HMM}}_{t,i}}{\sum\limits_{k=1}^K\alpha^{\tp{HMM}}_{t,k}\beta^{\tp{HMM}}_{t,k}},
   & \zeta^{\tp{HMM}}_{t,i,j}
  &=
  \frac{\alpha^{\tp{HMM}}_{t,i} \; a_{i,j} \; \mathcal{N}\big(\mb{\xi}_{t+1}|\;\mb{\mu}_j,\mb{\Sigma}_j\big) \; \beta^{\tp{HMM}}_{t+1,j}}
  {\sum\limits_{k=1}^K\sum\limits_{l=1}^K\alpha^{\tp{HMM}}_{t,k} \; a_{k,l} \; \mathcal{N}\big(\mb{\xi}_{t+1}|\;\mb{\mu}_l,\mb{\Sigma}_l\big) \; \beta^{\tp{HMM}}_{t+1,l}}.\label{eq: HMM_vars}
\end{align}
The expected complete log-likelihood of HMMs for a set of $M$ demonstrations, $\mathcal{Q}(\theta,\theta^{\; \text{old}}) =  \mathbb{E} \left \{ \sum_{m=1}^{M} \sum_{t=1}^{T} \log \mathcal{P}(\mb{\xi}_{m,t}, z_t|\theta) \; \big | \; \mb{\xi}, \theta^{\;  \text{old}} \right \}$, is maximized in an EM manner with
\begin{multline}
\mathcal{Q}(\theta, \theta^{\; \text{old}}) = \sum_{i=1}^{K} \sum_{m=1}^{M} \gamma^{\tp{HMM}}_{m,1,i} \log \Pi_i \; + \; \sum_{i=1}^{K} \sum_{j=1}^{K} \sum_{m=1}^{M} \sum_{t=1}^{T} \zeta^{\tp{HMM}}_{m,t,i,j} \log a_{i,j} \; + \\ \sum_{m=1}^{M} \sum_{t=1}^{T} \sum_{i=1}^{K} \mathcal{P}(z_t = i | \mb{\xi}_{m,t}, \theta^{\text{old}}) \log \mathcal{N}(\mb{\xi}_{m,t}| \mb{\mu_i, \mb{\Sigma_i}}).
\end{multline}
\begin{align*}
\emph{E-step:} \quad & \gamma^{\tp{HMM}}_{m,t,i} = \frac{\alpha^{\tp{HMM}}_{t,i}\beta^{\tp{HMM}}_{t,i}}{\sum\limits_{k=1}^K\alpha^{\tp{HMM}}_{t,k}\beta^{\tp{HMM}}_{t,k}},  & \\
\emph{M-step:} \quad & 	\Pi_i \leftarrow \frac{\sum_{m=1}^M \gamma^{\tp{HMM}}_{m,1,i} }{M} & 	a_{i,j} & \leftarrow \frac{\sum_{m=1}^M\sum_{t=1}^{T_m\!-\!1} \; \zeta^{\tp{HMM}}_{m,t,i,j}}
	{\sum_{m=1}^M\sum_{t=1}^{T_m\!-\!1} \; \gamma^{\tp{HMM}}_{m,t,i}}, \\
	\qquad & \mb{\mu}_i \leftarrow \frac{\sum_{m=1}^M\sum_{t=1}^{T_m} \; \gamma^{\tp{HMM}}_{m,t,i} \; \mb{\xi}_{m,t}}
	{\sum_{m=1}^M\sum_{t=1}^{T_m} \; \gamma^{\tp{HMM}}_{m,t,i} }
	,
	& \mb{\Sigma}_i &\leftarrow \frac{\sum_{m=1}^M\sum_{t=1}^{T_m} \; \gamma^{\tp{HMM}}_{m,t,i}
	(\mb{\xi}_{m,t}-\mb{\mu}_i) (\mb{\xi}_{m,t}-\mb{\mu}_i)^\trsp}
	{\sum_{m=1}^M\sum_{t=1}^{T_m} \; \gamma^{\tp{HMM}}_{m,t,i} } .
\end{align*}
Note that numerical underflow issues occur with a naive implementation of the above algorithm. In practice, a simple approach to avoid this issue is to rely on scaling factors during the computation of the forward and backward variables, which get canceled out when normalizing the posterior \cite{Rabiner89}. 
\subsection{Linear Quadratic Tracking}\label{app: LQT_deriv}
The discrete-time dynamical system for the double integrator is defined as,
\begin{equation}
  \overbrace{\begin{bmatrix}\mb{x}_{t+1}\\\mb{x}_{t+2} \end{bmatrix}}^{\mb{\xi}_{t+1}}
  =
  \overbrace{\begin{bmatrix}\mb{I}&\mb{\Delta} t\\\mb{0}&\mb{I} \end{bmatrix}}^{\mb{A}_d}
  \overbrace{\begin{bmatrix}\mb{x}_t \\ \mb{x}_{t+1}\end{bmatrix}}^{\mb{\xi}_t}
  +
  \overbrace{\begin{bmatrix} \mb{I} \frac{1}{2}\Delta t^{2} \\\mb{I} \Delta t \end{bmatrix}}^ {\mb{B}_d}
  \mb{u}_t.
  \label{eq:ABds}
\end{equation} The control law $\mb{u}_t^{*}$ that minimizes the cost function in Eq. \eqref{Eq: CostLQT} under \textbf{finite horizon} subject to the linear dynamics in discrete time is given as,
\begin{eqnarray} \label{Eq: discControlLaw}
\mb{u}_t^{*} &=&  -\left(\mb{R} + \mb{B}_d^{\trsp} \mb{P}_t \mb{B}_d\right)^{-1} \mb{B}_d^{\trsp}\mb{P}_t \mb{A}_d \left(\mb{\xi}_t - \mb{\hat{\mu}}_t \right) -  \left(\mb{R} + \mb{B}_d^{\trsp} \mb{P}_t \mb{B}_d\right)^{-1}  \mb{B}_d^{\trsp} \left( \mb{P}_t \left(\mb{A}_d \mb{\hat{\mu}}_t - \mb{\hat{\mu}}_t \right)  + \mb{d}_t \right), \nonumber \\ 
 & = & \mb{K}^{\ty{P}}_t (\mb{\hat{\mu}}_t^{x} - \mb{x}_t) + \mb{K}^{\ty{V}}_t (\mb{\hat{\mu}}_t^{\dot{x}} - \mb{\dot{x}}_t) - \left(\mb{R} + \mb{B}_d^{\trsp} \mb{P}_t \mb{B}_d\right)^{-1}  \mb{B}_d^{\trsp} \left( \mb{P}_t \left(\mb{A}_d \mb{\hat{\mu}}_t - \mb{\hat{\mu}}_t \right)  + \mb{d}_t \right),
\end{eqnarray} where $[\mb{K}^{\ty{P}}_t, \mb{K}^{\ty{V}}_t] = -\left(\mb{R} + \mb{B}_d^{\trsp} \mb{P}_t \mb{B}_d\right)^{-1} \mb{B}_d^{\trsp}\mb{P}_t \mb{A}_d$ are the full stiffness and damping matrices for the feedback term, and $\left(\mb{R} + \mb{B}_d^{\trsp} \mb{P}_t \mb{B}_d\right)^{-1}  \mb{B}_d^{\trsp} \left( \mb{P}_t \left(\mb{A}_d \mb{\hat{\mu}}_t - \mb{\hat{\mu}}_t \right)  + \mb{d}_t \right)$ is the feedforward term. $\mb{P}_t$ and $\mb{d}_t$ are respectively obtained by solving the Riccati differential equation and linear differential equation backwards in discrete time from terminal conditions $\mb{P}_{T_p} = \mb{Q}_{T_p}$ and $\mb{d}_{T_p} = \mb{0}$,
\begin{eqnarray}\label{Eq: DiscreteRDE}
\mb{P}_{t-1} &=&  \mb{Q}_t - \mb{A}_d^{\trsp} \Big(\mb{P}_t\mb{B}_d \left(\mb{R} + \mb{B}_d^{\trsp} \mb{P}_t \mb{B}_d\right)^{-1}\mb{B}_d^{\trsp} \mb{P}_t - \mb{P}_t \Big)\mb{A}_d, \\
\mb{d}_{t-1} &=&  \Big(\mb{A}_d^{\trsp}  - \mb{A}_d^{\trsp} \mb{P}_t\mb{B}_d \left(\mb{R} + \mb{B}_d^{\trsp} \mb{P}_t \mb{B}_d\right)^{-1}\mb{B}_d^{\trsp}\Big) \Big(\mb{P}_t \left(\mb{A}_d \mb{\hat{\mu}}_t - \mb{\hat{\mu}}_{t+1} \right)  + \mb{d}_t\Big).
\end{eqnarray} 

For the \textbf{infinite horizon} case with $T \rightarrow \infty$ and the desired pose $\mb{\hat{\mu}}_t = \mb{\hat{\mu}}_{t_0}$, the control law in \eqref{Eq: discControlLaw} remains the same except the feedforward term is set to zero and $\mb{P}_{t-1} = \mb{P}_{t} = \mb{P}$ is the steady-state solution obtained by eigen value decomposition of the discrete algebraic Riccati equation (DARE) in \eqref{Eq: DiscreteRDE} \cite{borrelli11}. To solve DARE, we define the symplectic matrix, 
\begin{equation}
\mb{H}_b = \begin{bmatrix} \mb{A}_d + \mb{B}_d \mb{R}^{-1} \mb{B}_d^{\trsp} (\mb{A}_d^{-1})^{\trsp} \mb{Q} & \mb{B}_d\mb{R}^{-1} \mb{B}_d^{\trsp} (\mb{A}_d^{-1})^{\trsp} \\ - (\mb{A}_d^{-1})^{\trsp} \mb{Q} & (\mb{A}_d^{-1})^{\trsp}\end{bmatrix}.
\end{equation} The eigenvectors of $\mb{H}_b$ corresponding to eigenvalues lying inside the unit circle are used to solve DARE. Let $\begin{bmatrix}\mb{V}_1^{\trsp} & \mb{V}_{21}^{\trsp}\end{bmatrix}^{\trsp}$ denote the corresponding subspace of $\mb{H}_b$, then the solution of DARE is, $\mb{P} = \mb{V}_{21} \mb{V}_1^{-1}$ and the control law takes the form,
\begin{equation}
\mb{u}_t^{*} = - (\mb{R} + \mb{B}_d^{\trsp} \mb{P} \mb{B}_d)^{-1} \mb{B}_d^{\trsp} \mb{P} \mb{A}_d (\mb{\xi}_t - \mb{\hat{\mu}_t}).
\end{equation} Both discrete and continuous time linear quadratic regulator/tracker can be used to follow the desired pose/trajectory. The discrete time formulation, however, gives numerically stable results for a wide range of values of $\mb{R}$. 
\subsection{Distance to Cluster Subspace vs Distance to Cluster Mean}\label{app: dist_subspace}
The distance of a datapoint $\mb{\xi}_t$ to an existing cluster with mean $\mb{\mu}_i$ is represented as: $\| \mb{\xi}_{t} - \mb{\mu}_{i} \|_2^{2}$. In contrast, we define the distance of a datapoint from the subspace of a cluster, $\text{dist}(\mb{\xi}_{t}, \mb{\mu}_{i} , \mb{U}_{i}^{d_{i}})^{2}$, as the difference between the mean-centered datapoint and the mean-centered datapoint projected upon the subspace $\mb{U}_{i}^{d_{i}} \in \mathbb{R}^{D \times d_{i}}$ spanned by the $d_{i}$ unit eigenvectors of the covariance matrix, i.e.,
\begin{equation}
		\text{dist}(\mb{\xi}_{t}, \mb{\mu}_{i} , \mb{U}_{i}^{d_{i}}) = 
	{\Big\|(\mb{\xi}_{t} - \mb{\mu}_{i}) - 
	\rho_i \mb{U}_{i}^{d_{i}}\mb{U}_{i}^{{d_{i}}^\trsp} (\mb{\xi}_{t} - \mb{\mu}_{i}) \Big\|}_2, 
	\label{Eq: distSS}
\end{equation} 
where
\begin{equation*}
	\rho_i = \exp\left(-\frac{\|\mb{\xi}_{t} - \mb{\mu}_{i}\|_2^{2}}{ b_m}\right)
\end{equation*}
weighs the projected mean-centered datapoint according to the distance of the datapoint from the cluster center $(0 < \rho_i \leq 1)$. Its effect is controlled by the bandwidth parameter $b_m$. If $b_m$ is large, then the far away clusters have a greater influence; otherwise nearby clusters are favored. Note that $\rho_i$ assigns more weight to the projected mean-centered datapoint for the nearby clusters than the distant clusters to limit the size of the cluster/subspace. Our subspace distance formulation is different from \cite{Wang15} as we weigh the subspace of the nearby clusters more than the distant clusters. This allows us to avoid clustering all the datapoints in the same subspace (near or far) together.
%
\subsection{Pick-and-Place with Obstacle Avoidance Results}
\begin{figure*}[!hbp]
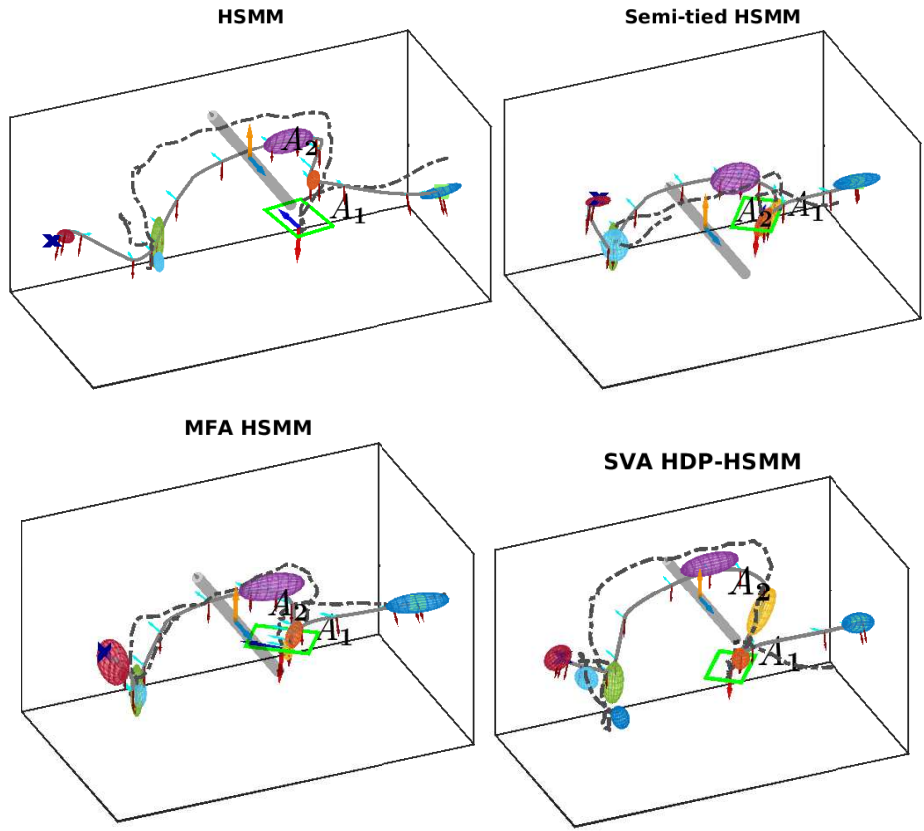

\centering
\includegraphics[trim={4.3cm 0.70cm 4.3cm 0.1cm},clip,scale = 0.65]{HSMM_PPv3.eps}
\includegraphics[trim={4.9cm 0.70cm 4.9cm 0.1cm},clip,scale = 0.65]{Semi_tied_HSMM_PP.eps}\\
\includegraphics[trim={4.9cm 0.70cm 4.9cm 0.1cm},clip,scale = 0.69]{MFA_HSMM_d1.eps}\;
\includegraphics[trim={5.1cm 0.70cm 5.2cm 0.1cm},clip,scale = 0.69]{sva_hdp_hsmm_1.eps}
\caption{\small Latent space representations of invariant task-parameterized HSMM for a randomly chosen demonstration from the test set. Black dotted lines show human demonstration, while grey line shows the reproduction from the model.} \label{fig: Baxter_PP_results_LS_Zoomed}
\end{figure*}
\end{document}